\documentclass[10pt,twocolumn,letterpaper]{article}
\usepackage{cvpr}
\usepackage{amsmath,amssymb,amsfonts}
\usepackage{algorithmic}
\usepackage{graphicx}
\usepackage{textcomp}

\usepackage{subcaption}
\usepackage{mathtools}

\newcommand{\secRef}[1]{Sec.\nolinebreak ~\ref{#1}}

\newcommand{\vecb}[1]{\boldsymbol{#1}}

\begin{document}
\title{BOSS: Bones, Organs and Skin Shape Model}
\author{%
{Karthik Shetty$^{1,2}$} ~ {Annette Birkhold$^2$}  ~ {Srikrishna Jaganathan$^{1,2}$}  {Norbert Strobel$^2$}  {Bernhard Egger$^1$} \and ~  {Markus Kowarschik$^2$} ~  {Andreas Maier$^1$} ~  \\
\normalsize $^1$FAU~Erlangen-Nürnberg, Erlangen, Germany \quad
\normalsize $^2$Siemens~Healthineers~AG, Forchheim, Germany\\
{\tt\small karthik.shetty@fau.de}
}
\maketitle

\begin{abstract}
Objective: A digital twin of a patient can be a valuable tool for enhancing clinical tasks such as workflow automation, patient-specific X-ray dose optimization, markerless tracking, positioning, and navigation assistance in image-guided interventions. However, it is crucial that the patient's surface and internal organs are of high quality for any pose and shape estimates. At present, the majority of statistical shape models (SSMs) are restricted to a small number of organs or bones or do not adequately represent the general population. Method: To address this, we propose a deformable human shape and pose model that combines skin, internal organs, and bones, learned from CT images. By modeling the statistical variations in a pose-normalized space using probabilistic PCA while also preserving joint kinematics, our approach offers a holistic representation of the body that can benefit various medical applications. Results: We assessed our model's performance on a registered dataset, utilizing the unified shape space, and noted an average error of 3.6 mm for bones and 8.8 mm for organs. To further verify our findings, we conducted additional tests on publicly available datasets with multi-part segmentations, which confirmed the effectiveness of our model. Conclusion: This works shows that anatomically parameterized statistical shape models can be created accurately and in a computationally efficient manner. Significance: The proposed approach enables the construction of shape models that can be directly applied to various medical applications, including biomechanics and reconstruction.

\end{abstract}

\section{Introduction}
\label{sec:introduction}
A virtual patient model that incorporates anatomy and physiology has the potential to enhance numerous medical diagnosis and therapy tasks. In particular, in the field of minimally invasive image-guided procedures, a detailed understanding of the patient being treated could facilitate (semi-) automated treatments. Given that not all necessary information about a patient is always available, a statistical model of body shape can serve as a foundation for incorporating patient-specific information. Such a Statistical Shape Model (SSM) of the human anatomy typically represents the average shape of multiple subjects and their variation in shape using a low-dimensional parameter space~\cite{smpl,scape}. Further, by incorporating kinematics, thereby the pose, the model can realistically represent a human body during a medical intervention.

The availability of such patient models opens up  possibilities for automating clinical workflow steps  entirely based on simulations. For example, in interventional environments such a model can be used to estimate and reduce the X-ray dose distribution within the patient as well as the dose the staff is exposed to due to scattered radiation~\cite{dose_estimation_1,dose_estimation_2}. Other applications are the automated positioning of the imaging system relative to a target region, such as an organ, or the virtual display of X-ray images that would be acquired based on the current system settings and position and patient pose prior to any X-ray exposure. It was also demonstrated such a generation of virtual X-ray images for bony structures based on pre-existing 3D hip models which are later iteratively fit to a patient's X-ray~\cite{hip_recon, maier2012fast}.

A virtual twin of the patient could be further employed for enhanced segmentation of anatomical structures~\cite{SSM_seg_1}, if (preoperative) images of a patient are available or are generated during the intervention. This would further facilitate the (pre- or intraoperative) generation of a patient-specific hierarchical model. It has been shown, that  coupling a statistical knee model with a segmentation neural network a more precise segmentation of the knee based on magnetic resonance images can be achieved~\cite{SSM_seg_2}. Shape models can be beneficial for the segmentation of 3D volumetric images such as Computed Tomography (CT) or Magnetic Resonance Imaging (MRI)~\cite{shape_model_seg1,shape_model_seg2,zhong2021deep}, as well as for 2D projection images such as X-rays~\cite{shape_model_seg3}. This is because the prior knowledge of the body shape can be used as a regularizer. Employed on a full body patient model, this technique could be applied online during image-guided (MR or X-ray) medical interventions and might therefore facilitate navigation during the procedure. Additionally, shape models can be used to generate synthetic data to train deep neural networks, allowing them to better generalize~\cite{shape_model_syn1,shape_model_syn2} to new subjects and overcome the limited anatomical variation often observed during domain translation. 

Anatomic body models are similar to statistical shape models in that they are both computer-generated models of the human anatomy. However, anatomic body models are designed to provide a highly detailed representation of the anatomy of a single individual, while statistical shape models aim to capture the variation in shape that occurs across the population. Anatomic body models typically include different body parts, such as bones, muscles, organs, and vessels~\cite{opensim,zygote}. They can be used for a variety of purposes, including visualization of anatomy for educational purposes, understanding relationships between different structures, and simulations of medical procedures. Despite their level of detail, the focus on a single individual's anatomy may limit their practicality in real-world applications. We hold the view that by leveraging the abundance of data and the progress in deep learning, the gap between statistical models and anatomical models can be bridged through a data-driven approach.

While prior studies in shape modeling focused primarily on specific anatomical regions (few bones or organs), this research aims to develop a comprehensive model representing the entire human body. Numerous works have proposed various solutions to create personalized anatomical models, such as Meng et.al.'s realistic spine model, which was learned from partial scans~\cite{spine_model}. Kadlecek et.al. proposed personalized anatomical models for physics-based animation using only the surface 3D scans of skin~\cite{recon1}. However, the bone structures in these models are based on uniform scaling, which may not accurately represent the true structure. While the individual components of the BASH model provide accurate representations of kinematics and skin surface, combining them using linear interpolation may not fully capture the true anatomy. The most similar model to our work is likely the recent OSSO model, which combines a skin and bone model~\cite{osso}. OSSO is a data-driven model that infers the shape of the skeleton from a given human surface scan. However, it is important to note that the skeleton model in OSSO is learned solely from 2D data, even though the method is data-driven.

In the field of patient modelling, the most prominent approach describing whole body anatomies is based on 4D extended cardiac-torso (XCAT)~\cite{xcat1} employing nonuniform rational B-spline (NURBS).  The underlying model is based on male and female CT scan from the Visible Human dataset~\cite{visiblehuman} which were manually segmented on a fixed pose. The model was extended to incorporate cardiac and respiratory motions as a parametric function, based on cardiac- and respiratory-gated multislice CT data. To enhance the model's representation, it was further extended to handle CT scans from individuals of varying age, height, and body mass~\cite{xcat2, xcat3}. However, this process is computationally expensive since each segmented CT scan must be registered to a template, followed by large deformation diffeomorphic metric mapping between the initial XCAT model and the registered scan. Furthermore, the robustness of this method decreases when patients are in different poses.

Various techniques have been introduced to create realistic models of human skin for individuals using captured 3D range scans that modify a generic 3D skin template to fit the particular person. For instance, Allen et al. employed non-rigid template fitting to compute correspondences between human body shapes in similar poses~\cite{fit1}. This method has been extended to work for varying poses~\cite{fit2, reg2}. A dataset of varying shapes and poses can be utilized to create a data-driven model for building a human skin surface shape model that incorporates these variations~\cite{scape, sscape, smpl}. These models represent the variations in human shape identity using a PCA space, and variations due to pose using an articulated skeleton-based skinning approach.

As these methods have shown promising results for the human skin, in this paper we extend the existing method based on SMPL~\cite{smpl} to combine  skin, skeleton and internal organs as one parameterized full body model. We start by creating an articulated template mesh for skeleton and organs. Followed by registration of the individual structures, which is then pose-normalized to a rest pose. The shape space is then learnt on the rest pose. This is the first joint model for bones, organs and skin covering the full body. This demonstrates the applicability of the concepts of statistical shape models to a full-body scale, encompassing multiple layers. Such a comprehensive model enables holistic image processing and we expect it to be highly valuable to improve image acquisition and understanding.

\section{Methodology}
Our statistical model is based on a set publicly available whole body as well as partial  CT scans (total~$\approx300$). We  performed automatic segmentation on them to isolate skeletons, internal organs and skin surfaces to model those components individually. Unlike SMPL or other well known statistical parametric models, the availability of a pose-only dataset is not feasible as we cannot subject a patient with unnecessary radiation. Usually while modelling the human skin, the pose-dataset is primarily used to learn the pose related shape variations, such as the bulging of the muscles.

\subsection{SMPL Introduction}
We provide necessary background on the SMPL framework, which forms the basis of our work~\cite{smpl}. The SMPL model is a statistical parametric function $M(\vecb{\beta}, \vecb{\theta}, \vecb{t}; \mathbf{\Phi})$, where $\vecb{\beta}$ are the shape parameters,  $\vecb{\theta}$ are the pose parameters, $\vecb{t}$ represents the global translation and, $\mathbf{{\Phi}}$ represents the learned model parameters. The output of this function is a triangulated surface mesh with $N=6890$ vertices. The shape parameters $\vecb{\beta} $ are represented as low-dimensional PCA coefficients, learned from a standard shape dataset~\cite{caesar} with $B_{S}(\vecb{\beta}): \mathbb{R}^{|\vecb{\beta}|} \mapsto \mathbb{R}^{3 N}$ representing offsets to a mean template mesh $\mathbf{\bar{T}}$ in the rest pose.

The pose of the model is defined by a kinematics chain with  a set of relative rotation matrices  $\mathcal{R}=\left[\mathbf{R}_{1}, \ldots, \mathbf{R}_{K}\right] \in \mathbb{R}^{3 \times 3}$ made up of $K = 24$ joints. Each set of rotation matrices is a function of the pose parameters $\vecb{\theta}_i \in \mathbb{R}^{3}$ which represents the axis-angle rotations relative to the joints. 
Deforming the template model from its rest pose to a desired pose based on the relative rotations is known as forward kinematics. Let $\vecb{t}_k \in \mathbb{R}^{3}$ represent the joint locations of the rest pose template. The new joint locations $\vecb{q}_k \in \mathbb{R}^{3}$ of the deformed model are determined by 

\begin{equation*}
\vecb{q}_k = \mathbf{G}_k(\vecb{t}_k - \vecb{t}_{p(k)}) + \vecb{q}_{p(k)},
\end{equation*}
where $\mathbf{G}_k {=}\mathbf{R}_k \mathbf{G}_{p(k)} \in \mathbb{R}^{3{\times}3}$ are the global rotation of each joint $k$ calculated recursively based on the kinematic tree. Here, $p(k)$ represents the parent joint for joint $k$. For more in depth description and implementations refer to~\cite{expmaps, smpl}.

The deformation process requires additional parameters such as the joint regressor $\mathcal{J}(.)$, blend weights $\mathcal{W}$ and pose-dependent shape variations $B_{P}(.)$. The joint regressor $\mathcal{J}(\vecb{\beta})$ computes the new joint location $\vecb{t}_k$ for different body shapes in the rest pose. The blend weights and pose-dependent shape variations are learned from a pose dataset consisting of multiple subjects in various poses.
The smoothing function of rotating vertices around a given joint is determined by the blend weights $\mathcal{W}$, using a given blend skinning function $W(.)$. The pose-dependent shape variations $B_{P}(\vecb{\theta}): \mathbb{R}^{|\vecb{\theta}|} \mapsto \mathbb{R}^{3 N}$ represents the offsets to a template mesh in the rest pose. Provided a template mesh  $\mathbf{\bar{T}}$, a morphed model can be represented as
\begin{equation}
\mathcal{M} =  W(\mathbf{\bar{T}} + B_{S}(\vecb{\beta}) +   B_{P}(\vecb{\theta}), \vecb{\theta}, \mathcal{J}(\vecb{\beta}), \mathcal{W}).
\end{equation}

In order to achieve smooth deformations, the Linear Blend Skinning (LBS) $W(.)$ method is used. This works by assigning blend weights $\mathcal{W}$ to each vertex of the mesh, which determine how much each segment of the skeleton affects the vertex's rotation. The resulting transformed vertex position $\mathbf{v}{i}^{\prime}$ is given by the sum of the contributions from each segment
\begin{align*}
\mathbf{v}_{i}^{\prime}=\sum_{k=1}^{K} w_{k, i} \mathcal{G}_{k} \mathbf{v}_{i}
\end{align*}
Here, $w_{k, i}$ is an element of the blend weights $\mathcal{W}$ that corresponds to segment $k$ and vertex index $i$, and $\mathcal{G}_{k} {=} [\mathbf{G}_k|\vecb{q}_k]$ is the global transformation of joint $k$.

\subsection{Data}\label{sec_data}
In this study, 3D CT images of nearly 300 patients were used for the statistical modeling. Due to the limited public availability of whole body CT scans, we relied on two types of datasets, whole-body scans and partial-body scans. Whole-body scans typically consists of the entire human body from head to toe with a exceptions missing parts around the arm regions. In total, 42 valid scans from the Visceral dataset~\cite{visceral} make up the whole-body scans with an average voxel resolution of $0.87 \times 0.87 \times 2.0$~mm. The partial-body scans, on the other hand, predominantly cover scan areas ranging from the neck to the femur region. This set comprises 58 valid scans from the Visceral dataset~\cite{visceral}, and 206 valid scans from the QIN-HeadNeck dataset~\cite{qin} with an average voxel resolution of $0.91 \times 0.91 \times 1.5$~mm and $0.97 \times 0.97 \times2.0$~mm, respectively. Additional metadata such as the height and weight of the patient during the CT acquisition was available for the QIN-HeadNeck dataset. On both datasets we have only included patients who have 12 thoracic vertebra and 5 lumbar vertebra as we are interested in the general population who can modeled based off the template skeleton model. Hence, we cannot account for varying amount of bones. In addition we excluded all patients with significant metallic implants as they cause imaging artifacts which degrade the quality of the segmented CT images. We further rejected patients with missing organs such as the kidneys. In total, we make use of 306 CT scans for building the model.

\begin{figure}[htb]
     \centering
     \begin{subfigure}[b]{0.2\textwidth}
         \centering
         \includegraphics[width=\textwidth,keepaspectratio]{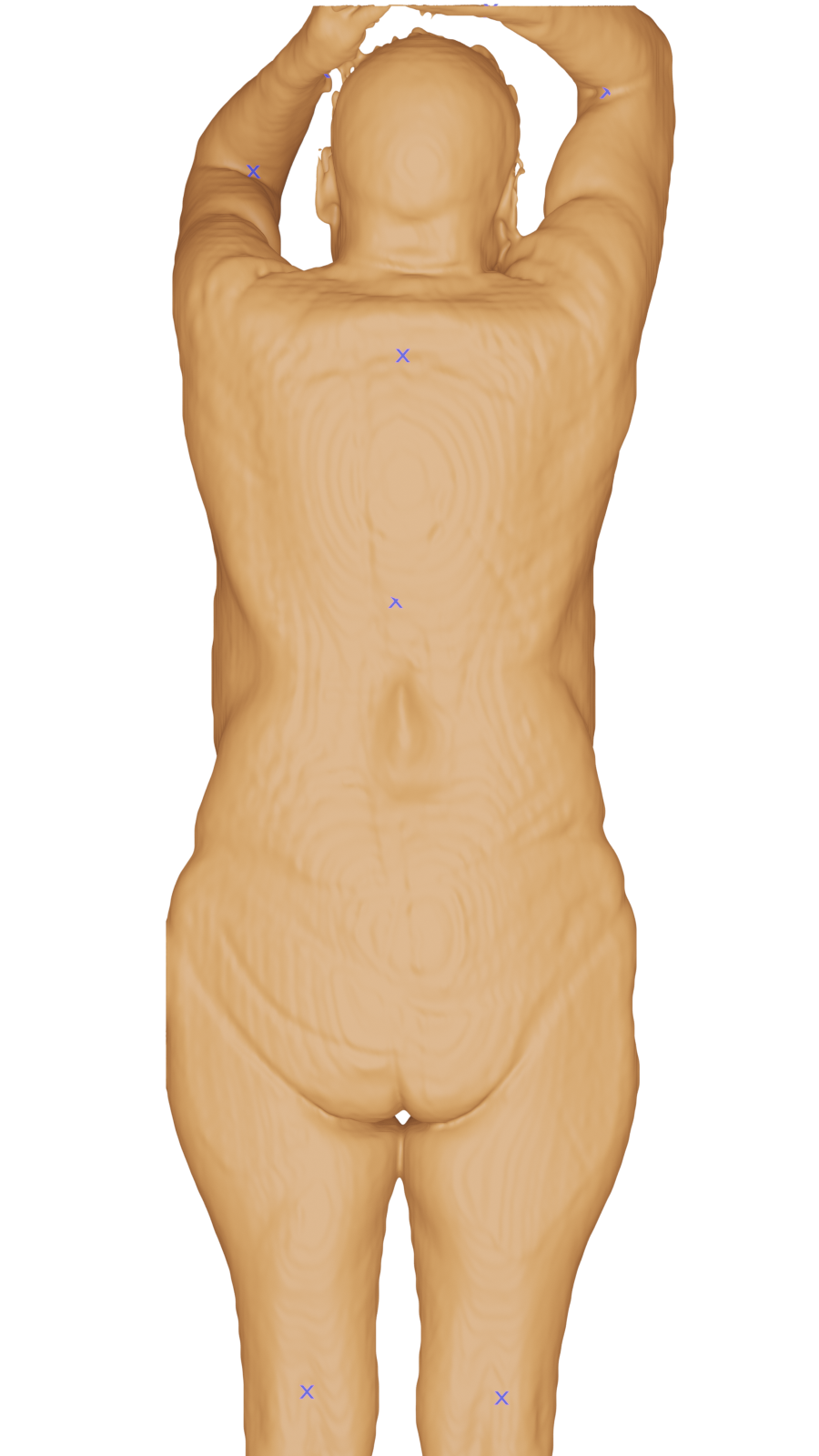}
         \caption{}
     \end{subfigure}
     \begin{subfigure}[b]{0.2\textwidth}
         \centering
         \includegraphics[width=\textwidth,keepaspectratio]{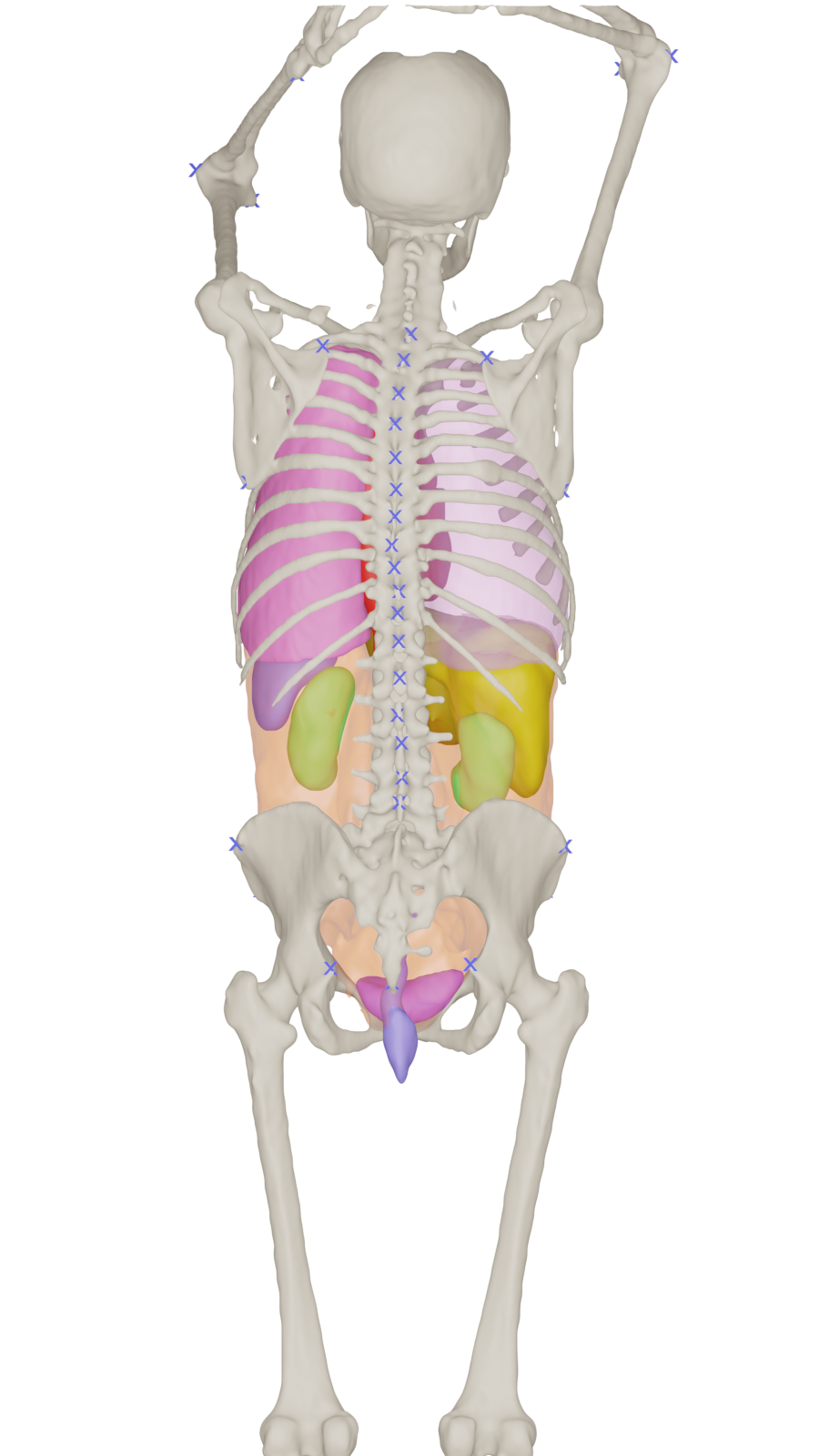}
         \caption{}
     \end{subfigure}

    \caption{Example surface mesh with landmarks (depicted in the image with a blue cross) from the Visceral dataset~\cite{visceral} of (a) skin and (b) bone-organ obtained from a segmented CT scan.}
    \label{fig:segmentation}
\end{figure}

We anatomically segmented the 3D scans into 3 distinct sets, namely skin, bones and organs. This was automatically performed using the AI-Rad Companion Organs RT1 software (Siemens Healthcare GmbH, Erlangen, Germany). Note that we use the term organs loosely in this context as it refers to a combined representation of lungs, liver, heart, kidneys, bladder, rectum, esophagus, aorta, and the bowel region. From the segmented volumes we extracted skin $\mathbf{S}_s^i$, bone $\mathbf{S}_b^i$, and organ $\mathbf{S}^{i,k}_o$ surface meshes using the Marching Cubes~\cite{marchingcubes} algorithm. Here $i \in [1,|\mathbf{S}|]$ represents the volume index and $k$ represents the organ index. Fig.~\ref{fig:segmentation} shows an example surface mesh of a patient. Further, we manually annotated around 60 landmarks $L^i_j$ both on the skin and on the bones with the bone's structure as reference guides. This also acts as a guide in determining the availability of segments such as the arms and legs.

\subsection{Skin Model}\label{sec_reg}
In order to create a consolidated human model, it is fundamental that we establish dense correspondence between all scans in the dataset. The typical approach of achieving this is to start from a common template mesh which ensures identical mesh topology for all the scans~\cite{reg1,reg2}. This is accompanied by a skeleton-based deformation of the template to estimate the rough pose and shape of the scan. Subsequently a non-rigid deformation technique is tasked to fit the template to the surface scan. The two stage process of registering reduces the convergence speed and improves accuracy drastically. To avoid falling onto a local minima we make use of previously described manual landmarks to achieve the initial pose deformation of the template.

\subsubsection{Skin Registration}
With the availability of a trained SMPL model, the registration process is simplified due to the preexistence of a SSM. Hence, the pose and shape can be optimized simultaneously by minimizing  $E_{lm} + E_{data}$, where $E_{lm}$ is the landmark-loss  (Eq.~\ref{eq:e_l_skin}) and $E_{data}$ the data-loss  (Eq.~\ref{eq:e_data_skin}). 

\begin{equation}\label{eq:e_data_skin}
 E_{data} = \lambda_{d1} E_{d}(\mathbf{S}_s^i, M_s) + \lambda_{d2} E_{d}(M_s, \mathbf{S}_s^i)
\end{equation}

\begin{equation}\label{eq:e_dist_skin}
E_d(\mathbf{S}, M(\vecb{\beta}, \vecb{\theta}, \vecb{t}; \mathbf{\Phi})) = \sum_{m_{j} \in (M) }p_j\rho\left(||m_j - \mathcal{N}(\mathbf{S})||_2^2  \right )
\end{equation}

\begin{equation}\label{eq:e_l_skin}
E_{lm} = \sum_{l_{j} \in L(M)} \left\| l_{j} - L_j \right\|_1
\end{equation}

The data loss accounts for the distance between the skin SMPL model $M$ and the surface $\mathbf{S}_s^i$. As correspondence is not present implicitly, we select the nearest neighbour $\mathcal{N}$ in $\mathbf{S}_s^i$ for any given vertex in $M$  for $E_{d}(\mathbf{S}_s, M_s)$ and inversely for $ E_{d}(M_s, \mathbf{S}_s)$. For robustness, we discard matches where the angle between the corresponding normals are above a threshold of $30^{\circ}$ and when the distance between the points are greater than $30$~mm by setting $p_j$ in \eqref{eq:e_dist_skin} to either $1$ or $0$. We make use of the robust Geman-McClure function~\cite{Geman1987StatisticalMF}, represented by $\rho$ to handle noisy data. 
The data-loss alone could fit the two surfaces if they are close and and on comparable poses, else the optimization process may get stuck in a local minima. The landmark-loss penalizes misalignment between the set of manually annotated landmarks $L^i_j$ of each scan with the corresponding  vertices $L(M)$ of the skin model. In addition to $E_{lm}$  and $E_{data}$ we further add regularization in the form of pose $E_\theta$, shape $E_\beta$, weight $E_w$ and height prior $E_h$. The pose prior is the same as Eq. 5 from~\cite{keepitsmpl}. Un-natural poses especially for the arms could lead to lower data error, hence these terms tries to keep the poses in a realistic range. We make use of the Gaussian mixture model provided by~\cite{keepitsmpl} for the pose prior. As the patient are lying on a table in a supine position, whereas the SMPL model was trained on standing pose we further add an exponential pose loss along the sagital plane for the joints near the thorax and abdomen region. The shape prior $E_\beta = \sum\left\|\beta\right\|$ forces the shapes to be close to the mean shape. For the partial-body scans, we add additional prior loss $E_h$ and $E_w$, when the patients height and weight are known. The height and weight of the skin model are measured in the rest pose for a given shape parameter $\vecb{\beta}$. The height is measured from the head to toe, whereas the weight is measured as function of the mesh volume $V(M)$ as described in ~\cite{w2vol}. 

\begin{equation}\label{eq:e_skin_weight}
E_w = ||(V(M_s^i) + 4.937)/1.015 - w^i||_2^2 .
\end{equation}

We optimize the following energy term 

\begin{equation}\label{eq:skin_rigid}
\begin{split}
\underset{\vecb{\beta}, \vecb{\theta}, \vecb{t}}{\arg \min } ( E_{lm} {+} E_{data} {+} E_{\theta} {+} E_{\beta} {+} E_{w} {+} E_{h}),
\end{split}
\end{equation}
to obtain an initial fit $\mathbf{\Tilde{V}}^i_s$. The weight terms $\lambda_x$ associated with the energy term $E_x$ is omitted for easier readability. The initial fit is obtained by first by minimizing the global translation $\vecb{t}$, followed by the pose $\vecb{\theta}$, shape $\vecb{\beta}$ and finally all three parameters simultaneously.

If the scans $\mathbf{S}_s^i$ were in a standing pose, registration would be achieved under the assumption that the SMPL model represents an adequate space of human shape variations. However, the model we would like to represent are the ones in a supine position, taken during a CT procedure. This usually causes the backs to become flat, stomach to be depressed and the chest bulged out. Hence, we perform non-rigid registration on the initial fit $\mathbf{\Tilde{V}}^i_s$. Similar to the works from~\cite{fit1, fit3}, we represent a set of $3 \times 4$ affine transformation matrices $\mathbf{A}_j^i$ associated with each vertex of the initial fit $\mathbf{\Tilde{V}}^i_{s,k}$, with the aim to align the vertices to the scan $\mathbf{S}_s^i$. This is achieved by minimizing $E_{lm} + E_{data} + E_s + E_o$, where $E_s$ and $E_o$ are smoothing and orthogonality constraints~\cite{fit3} respectively. To achieve local rigidity, the affine transformations applied on the vertices $\mathbf{\Tilde{V}}^i_{s,k}$ need to be close to the transformations on the neighbouring vertices $\mathbf{\Tilde{V}}^i_{s,k} \in \mathcal{N}(\mathbf{\Tilde{V}}^i_{s,j})$. Therefore, the smoothness term $E_s(\mathbf{\Tilde{V}}^i_s)$  can be defined as

\begin{equation}\label{eq:e_s_skin}
E_s(\mathbf{p}) = \sum_{\mathclap{\left\{j, k \mid\left\{\mathbf{p}_{j}, \mathbf{p}_{k}\right\} \in \text {edges}(\mathbf{p})\right\}}}c_{ij}\left\| \mathbf{A}_{j}\mathbf{p}_{j}-\mathbf{A}_{k}\mathbf{p}_{k}\right\|^{2}_2.
\end{equation}
Here, $c_{ij}$ represents the Laplacian cotangent edge weights, which tries to make changes
on the transformation matrices $\mathbf{A}_j^i$ over the mesh as smooth as possible~\cite{cot}. The orthogonality constraint additionally preserves local rigidity during registration by enforcing the affine transformation to be close to a rigid transformation by

\begin{equation}\label{eq:e_o_skin}
\begin{aligned}
E_o=\sum_{j}\left\|\mathbf{A}_{j}^i -\mathbf{R}_{j}^i\right\|_{F}^{2}. \\
\end{aligned}
\end{equation}

Here $\mathbf{R}_{j}$ is the closest projection of $\mathbf{A}_{j}$ onto the rotation matrix group. This can be extracted by performing Singular Value Decomposition on the transformation matrix. All energies are minimized to obtain a final non-rigid fit $\mathbf{V}^i_{s}$ using a gradient-based LBFGS~\cite{bfgs} minimization method and make use of automatic differentiation packages.

The main advantage of the two step process of registration is that it can handle scans with missing data or holes. Missing data here refers to the non-availability of scans sections such as the arms or legs from the partial-body dataset. Missing data is identified by the non-availability of landmarks for a given scan. As the SMPL model is divided into 24 sections, we can prevent pose deformation and data loss minimization on those sections. Using a SSM reduces the search space, and can provide shape in the realm of probable human shapes.

\subsection{Bone-Organ Model}\label{bone_model}

\begin{figure}[htb]
    \centering
    \includegraphics[width=0.7\linewidth,keepaspectratio]{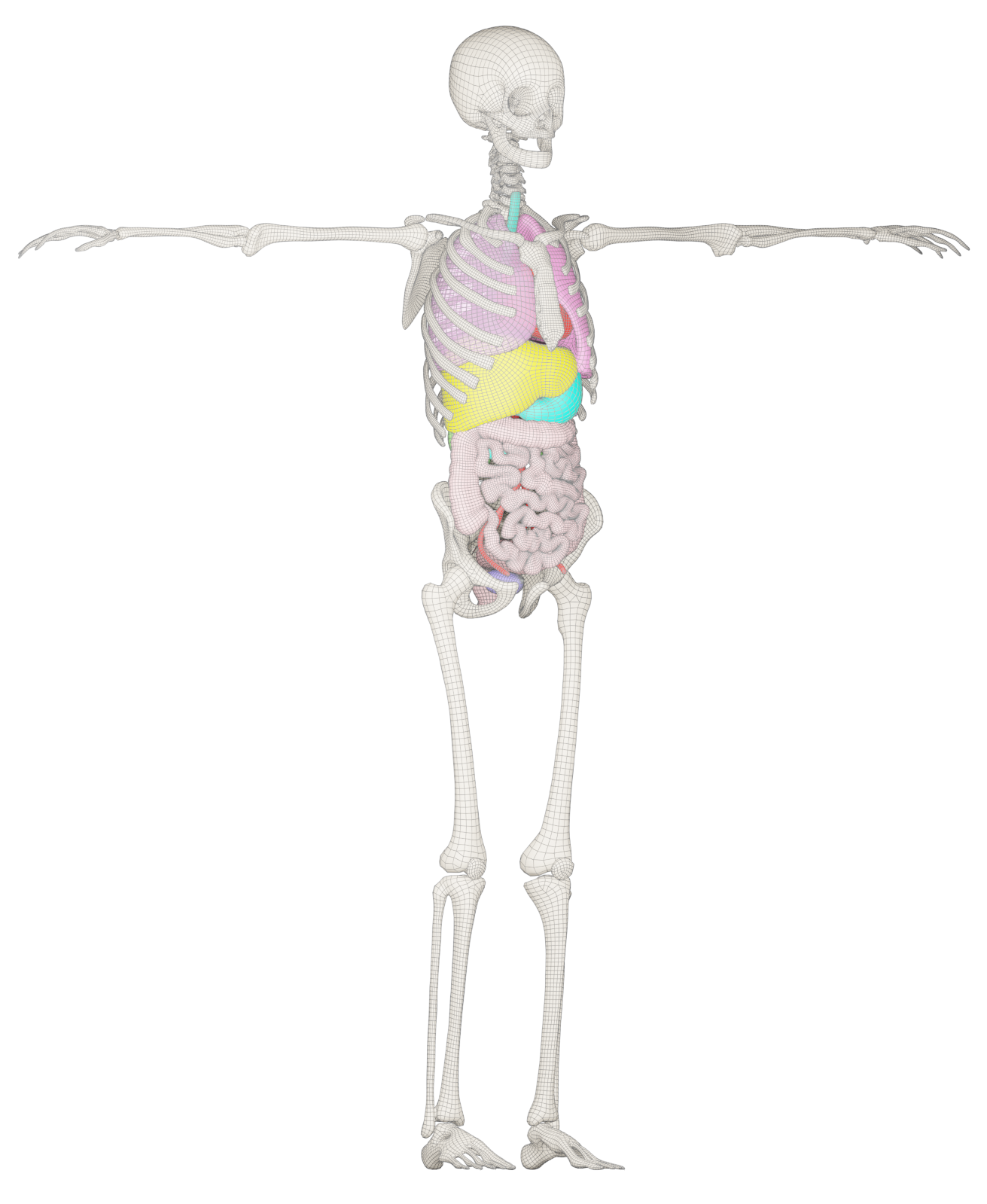}
     \caption{Template bone-organ model, consisting of lungs, liver, kidneys, spleen, heart along with aorta and bladder }
    \label{fig:bone_organ_model}
\end{figure}

Unlike the skin model, publically available SSMs for bones and organs do not exist. For this purpose we create a deformable model from scratch. A template mesh is derived from an existing polygon data BodyParts3D, which was extracted from full body MRI images~\cite{bodyparts3d}. The bone model is made up of $70$ segments, including skull, femur, humerus, forearm, lower leg, scapula, clavicle, sternum, hands, feet, vertebra, ribs, and pelvis. The organ model includes lungs, liver, kidneys, spleen, heart, bladder, rectum, esophagus, and aorta. We also incorporate the bowel region containing the stomach and intestines. However, segmentation for the individual bowel components are not available, rather a hull enclosing the stomach and intestines. The entire bone-organ template is made up of $104,546$ vertices and $209,418$ faces, of which $65,617$ vertices are made up of the bone section and the remaining for the internal organs.

On top of the template, we define a kinematics chain made up of $N_b = 63$ joints comprising $63$ segments. Though we start with $70$ individual segments, we consider femur-patella and all cervical vertebrae as combined segments. Linear blend skinning is adopted on femur-patella-tibia and cervical section to achieve a smooth deformable bone model. The blend weights are set to 1 for the rigid entity with respect to their own segments, whereas for the composite structure it is evenly  distributed between the parent and child segment. The initial blend weights for the organs are set only with respect to the vertebral section. We make use of Blender~\cite{blender} to automatically generate these weights.

We rigidly deform the vertebra of the bone-organ mesh model to one of the segmented CT volumes of comparable shape and size, such that the mesh represents a person laying in supine pose. Similarly, we define the skin template $\mathbf{\Tilde{T}}_{s}$ in supine pose by re-posing the non-rigid skin model to a T-pose of the same CT volume. Additionally, we also rotate the arms and legs of the bone-organ model, such that they lay inside and follow the same T-pose as the skin from the SMPL model. The final template bone-organ template mesh $\mathbf{\Tilde{T}}_{bo}$ is shown in Fig.~\ref{fig:bone_organ_model} .

\subsubsection{Bone Registration}\label{bone_reg}
The registration process in general follows the methodology as described in \secRef{sec_reg}. However, estimating the rough pose followed by non-rigid registration is not feasible by virtue of the complex thin structure of bones. This problem is particularly evident on the scapula and clavicle, which leads to incorrect poses for the rest of the template. To address this, we simultaneously estimate a rough shape and pose. The shape variations are achieved by applying a scale transform along a segment in  world coordinates. Consequently, the joint locations along the kinematic chain are also scaled by the same amount.

Hereby a simplistic deformable bone model can be expressed as $M^b(\vecb{\hat{\beta}}_{b}, \vecb{\theta}_b, \vecb{t}_b; \mathbf{\Phi}_b)$, where $\vecb{\hat{\beta}}_b \in \mathbb{R}^{63 \times 3}$ represents the scaling parameters, $\vecb{\theta}_b \in \mathbb{R}^{63 \times 3}$ represents the pose parameters, $\vecb{t}_b \in \mathbb{R}^{63 \times 3}$ represents the individual segment translation parameters,  and $\mathbf{\Phi}_b$ represent the model parameters comprising of the kinematic chain and the initial blend weights.

\begin{figure}[htb]
    \centering    \includegraphics[width=0.4\linewidth,keepaspectratio]{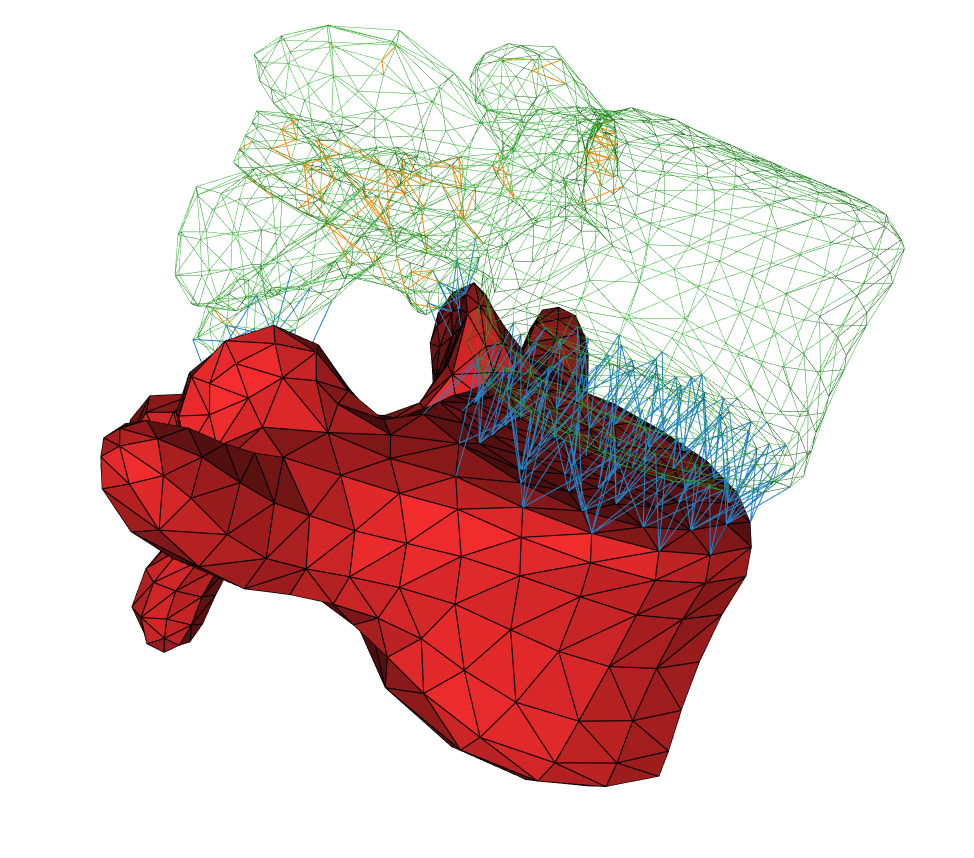}
      \caption{Example of the virtual edges between two vertebral segments to prevent mesh overlap during non-rigid registration. Blue and Orange lines represent inter-segment and intra-segment virtual edges respectively.  }
    \label{fig:fake_edge}
\end{figure}

Similar to the skin registration from \secRef{sec_reg}, registration of the bones can be performed with a few minor changes as described
\begin{equation}\label{eq:skel_rigid}
\begin{split}
\underset{\vecb{\hat{\beta}}_b, \vecb{\theta}_b, \vecb{t}_b}{\arg \min } ( E_{lm} {+} E_{data} {+} E_{\theta_b} {+} E_{\hat{\beta}_b} {+} E_{t_b} {+} E_{lm}^s).
\end{split}
\end{equation}

The bone surface scans, represented by $\mathbf{S}_b^i$, are made up of 26 separate segments including 24 ribs, the sternum, and the rest of the skeletal structure. The set of vertices corresponding to a respective segment on the surface scans and template mesh are represented by $\mathbf{ss}$ and $\mathbf{sm}$ respectively. The data loss $E_{data}$ term from Eq.~\ref{eq:e_l_skin} is replaced with Eq.~\ref{eq:e_d_bone}.
The pose $E_{\theta_b} {=} \sum||\vecb{\theta}_b^j {-} \vecb{\theta}_b^{p(j)}||$ and shape $E_{\hat{\beta}_b} {=} \sum||\hat{\vecb{\beta}}_b^j {-} \hat{\vecb{\beta}}_b^{p(j)}||$ prior loss forces the registered model maintain the original template shape and prevent unnatural poses. Here, $p(j)$ denotes the parent joint of joint $j$. The pose prior is defined only for the vertebral section. As the shape here represents a scaling function, the shape prior forces the scaling to be similar to that of its parent segment. The incorporation of individual translation $\vecb{t}$ allows free floating segments such as the sternum, scapula, etc. to move freely. However, we add a translation prior $E_{t_b} {=} \sum||\vecb{t}_b^j||$ that prevents large translation movement far away from its initial position. In addition to the landmark error $E_{lm}$, we include a skin-based vicinity landmarks $E_{lm}^s$. This helps to generate an estimated fit of the bones in the case of missing data around the leg and arm regions. Here, we use a learnt mapping between the skin vertices and joints of skeletons $\mathcal{J}(M_s)$, in particular for the  arms, legs, hand and feet. The mapping is learnt on the set of registered scans where aforementioned sections where present in the CT data. Using the registered skin as a reference, we minimize the loss between the predicted joint locations $\mathcal{J}(M_s)$ and the joint locations obtained from the skeleton model $M^b(\vecb{\beta}^b, \vecb{\theta}^b)$.

\begin{equation}\label{eq:e_d_bone}
\begin{split}
 E_{data} = \sum_{\mathclap{\left\{\mathbf{sm}_k,\mathbf{ss}_k \in \text {segments}(M_b, \mathbf{S}_b)\right\}}}\left( \lambda_{d1} E_{d}(\mathbf{sm}_k, \mathbf{ss}_k) + \lambda_{d2} E_{d}(\mathbf{ss}_k,\mathbf{sm}_k) \right)
\end{split}
\end{equation}

Similar to the skin, non-rigid registration is performed for the bones by minimizing $E_{lm} + E_{data} + E_s + E_o$. Unlike skin, the bones are narrow structures, resulting in severe mesh overlap when using the same formulation. To prevent this, we create inter-segment and intra-segment virtual edges on the template mesh as shown in Fig.\ref{fig:fake_edge}. These virtual edges are defined based on a distance threshold and direction of the vertex normals. For inter-segment edges the normals between two set of vertices  need to be facing each other whereas for the intra-segment edges the normals need to be facing away from each other.

\begin{figure}[htb]
    \centering    \includegraphics[width=1\linewidth,keepaspectratio]{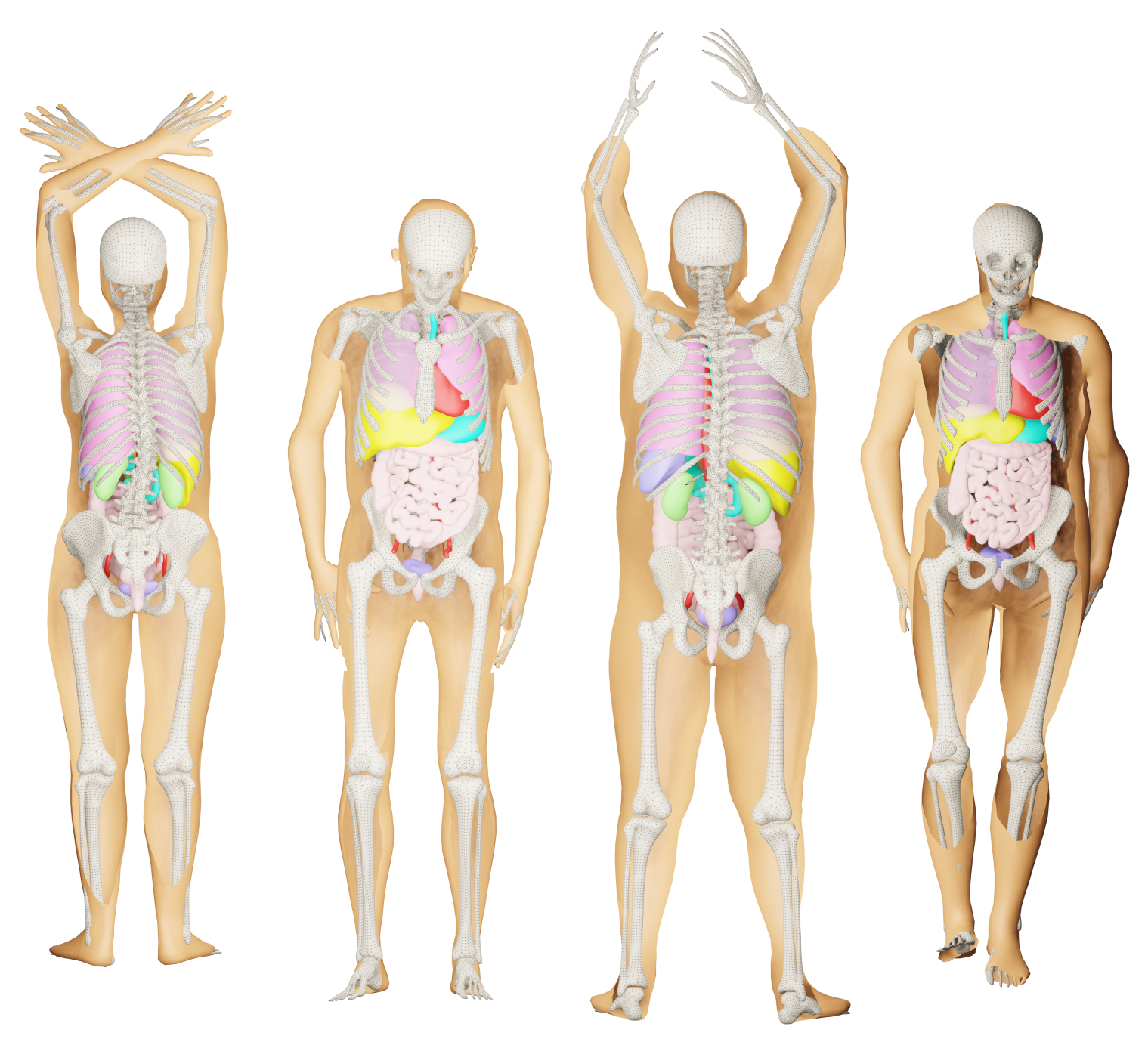}
      \caption{Examples of registered patients.  }
    \label{fig:registred_examples}
\end{figure}

Organ registration is a straightforward process that is similar to skin registration. The initial fit is obtained during the rigid registration of the bones. The non-rigid registration is accomplished through affine transformations applied directly to individual translations. To avoid mesh overlap, we include virtual edges between organs. However, the bowel region presents a challenge because there is no distinct segmentation, only a boundary. To address this, we use a higher regularizer loss and a penetration loss. The penetration loss calculates if any vertices of the bowel are inside another body region and penalizes the distance to the nearest boundary. Examples of registered scans are presented in Figure~\ref{fig:registred_examples}.

\subsection{Model Formulation}\label{model_form}
With the availability of registered scans  $\mathbf{V}^{s}$ and $\mathbf{V}^{bo}$ in the common mesh topology, we can now formulate the deformable shape model for the skin and internal body. It is necessary that all scans are normalized to the rest pose as defined by their respective templates $\mathbf{\Tilde{T}}^{s}$ and $\mathbf{\bar{T}}^{bo}$ in order to remove variance related to body articulation. We first perform the unposing (transforming to rest pose) operation on the skin, followed by the internal body. Unposing the skin first, allows us to use the skin as a guide for unposing the internal body later. From the pose-normalized models, it is then possible to learn a joint shape-space of the skin-bone-organ model.

\subsubsection{Skin Unposing}
We unpose the registered skin mesh $\mathbf{V}^{i}_s$ to the rest pose $\mathbf{U}^{i}_s$, defined by the template skin mesh $\mathbf{\Tilde{T}}_{s}$. As we prefer the patient skin model to operate similarly to that of the original SMPL model, we optimize for a new joint regressor $\mathcal{J}_s$, blend weights $\mathcal{W}_s$ and pose related shape variations $B_{P_s}$. This is done by minimizing the distance between the skin model vertices and the registration vertices as described in the following equation 

\begin{equation}\label{eq:model_params_1}
\underset{\mathbf{U}_{s}^i,  \vecb{\Tilde{\theta}}_{s}^i, B_{P_s},\mathcal{J}_s, }{\arg \min } \sum_{i}^{S_n}\left\|  W(\mathbf{U}_{s}^i {+} B_{P_s}(\vecb{\Tilde{\theta}}_{s}^i), \vecb{\Tilde{\theta}}_{s}^i; \mathcal{J}_s, \mathcal{W}_s) {-}\mathbf{V}^{i}_{s}\right\|^{2}.
\end{equation}

During optimization, we initialize $\mathcal{J}_s$,  $\mathcal{W}_s$ and $B_{P_s}$ with the original SMPL parameters. The rest pose vertices $\mathbf{U}^{i}_s$ are initialized  $M(\vecb{\beta}^i)$ with only the shape parameters $\vecb{\beta}^i$ obtained from the initial rigid skin fit. Similarly, the poses $\vecb{\Tilde{\theta}}_{s}^i$ are initialized with the poses $\vecb{\theta}^i$ obtained from the initial rigid skin fit. To stabilize the optimization, we make use of  multiple regularizers based on various assumptions.

\begin{figure*}[ht]
	\centering
	    \includegraphics[width=\textwidth,keepaspectratio]{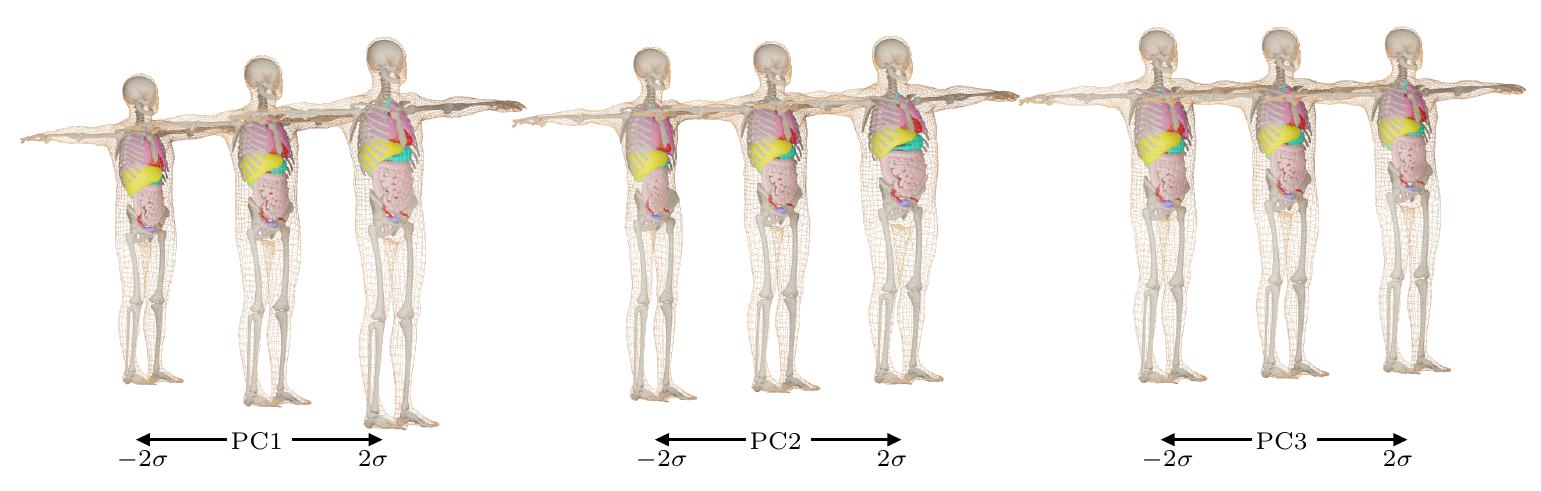}
 	\caption{The first three principal components of body shape are shown, varying about 2 standard deviations after normalizing the variance. One could infer that the height and weight of the patient are mostly explained by the first two components}\label{fig:pca}
\end{figure*}

As, the pose normalized subjects $\mathbf{U}^{i}_s$ needs to be aligned to the template mesh $\mathbf{\Tilde{T}}_{s}$, we add an edge loss of the form $\sum_{i}^{S_n}||\mathbf{U}^{i}_{s,e} {-} \mathbf{\Tilde{T}}_{s,e}||$, where $\mathbf{U}^{i}_{s,e}$,$\mathbf{\Tilde{T}}_{s,e} \in \text {edges} (\mathbf{\Tilde{T}}_s)$ represents the normalized direction vector for a  pair of neighbouring vertices. We further add constraints on $\vecb{\Tilde{\theta}}_{s}^i$, $\mathcal{J}_s$,  $\mathcal{W}_s$ and $B_{P_s}$ in the form of $L2$ loss, to not deviate too much from its initial values. In the original SMPL model, the joint regressor was computed using non-negative least squares, with a constraint that the weights add up to $1$. We maintain similar setting during the joint optimization by normalizing the joint regressor as $|\mathcal{J}_s|/|| |\mathcal{J}_s| ||_1$. During the optimization, we further add a regularizing term defined as
\begin{align*}
\sum_{i}^{S_n}||\mathcal{J}_s(W(\mathbf{U}_{s}^i,  \vecb{\Tilde{\theta}}_{s}^i, B_{P_s},\mathcal{J}_s)) - \mathcal{J}(\mathbf{\Tilde{V}}^i_s)||^2,
\end{align*}
on the joint locations, such that it is close to the joint locations from the original rigid fit.

Symmetry regularizer is applied on the pose-normalized mesh vertices $\mathbf{U}^{i}_s$ and on its joint locations $\mathcal{J}_s(\mathbf{U}^{i}_s)$ to enforce symmetry along the sagittal plane defined as
\begin{equation}
\sum_{i}^{S_n}||\mathbf{U}^{i}_{s} {-} \mathrm{m}(\mathbf{U}^{i}_{s})|| + ||\mathcal{J}_s(\mathbf{U}^{i}_s) {-} \mathrm{m}(\mathcal{J}_s(\mathbf{U}^{i}_s))||^2,
\end{equation}
where $\mathrm{m}$ denotes the mirror vertices or mirror joints. 

\subsubsection{Bone-Organ Unposing}
Similar to the skin model, we perform unposing of the bone-organ registered mesh $\mathbf{V}^{i}_{bo}$ to the rest pose $\mathbf{U}^{i}_{bo}$. However, performing exactly the same would lead to incorrect skeletal localization and mesh overlap of the internal body and the skin. 

The objective function defined in Eq.~\ref{eq:model_params_2} is to unpose the body-organ model by minimizing the distance between the model vertices $\mathbf{\bar{V}}_{bo}^i {=} W(\mathbf{U}_{bo}^i, \vecb{\Tilde{\theta}}_{bo}^i; \mathcal{J}_{bo}, \mathcal{W}_{bo})$ and the registration vertices. Additionally, the objective function defined in Eq.~\ref{eq:model_params_2} maintains the distance $\mathcal{P}$ between the skin and bone-organ model among both posed and unposed states as shown in Eq.~\ref{eq:model_params_3}. We do this with pose-only $\vecb{\bar{\theta}}_{bo}^i$ deformation of the unposed model $\mathbf{\bar{U}}_{bo}^i {=} W(\mathbf{U}_{bo}^i, \vecb{\bar{\theta}}_{bo}^i; \mathcal{J}_{bo}, \mathcal{W}_{bo})$.
For each skin vertex of the registered skin mesh $\mathbf{V}_{s}^i$, we find at most $6$ bone vertices from the registered bone-organ mesh $\mathbf{V}_{bo}^i$ based on empirically chosen distance (based on human size and the particular bone segment) and vertex normal  (skin and bone normal within $30\deg$) thresholds. 
However, we ignore skin vertices around the complex shoulders, elbows and knees joints, along with bone vertices for scapula and clavicle.  This avoids any pose related influences during optimization around these regions. 

\begin{equation}\label{eq:model_params_2}
\underset{\mathbf{U}_{bo}^i,  \vecb{\Tilde{\theta}}_{bo}^i,\mathcal{J}_{bo}, \mathcal{W}_{bo}}{\arg \min } \sum_{i}^{S_n}\left\|  \mathbf{\bar{V}}_{bo}^i {-}\mathbf{V}^{i}_{bo}\right\|^{2}; \\ 
\end{equation}

\begin{figure}[hb]
\centering
\includegraphics[width=0.8\linewidth]{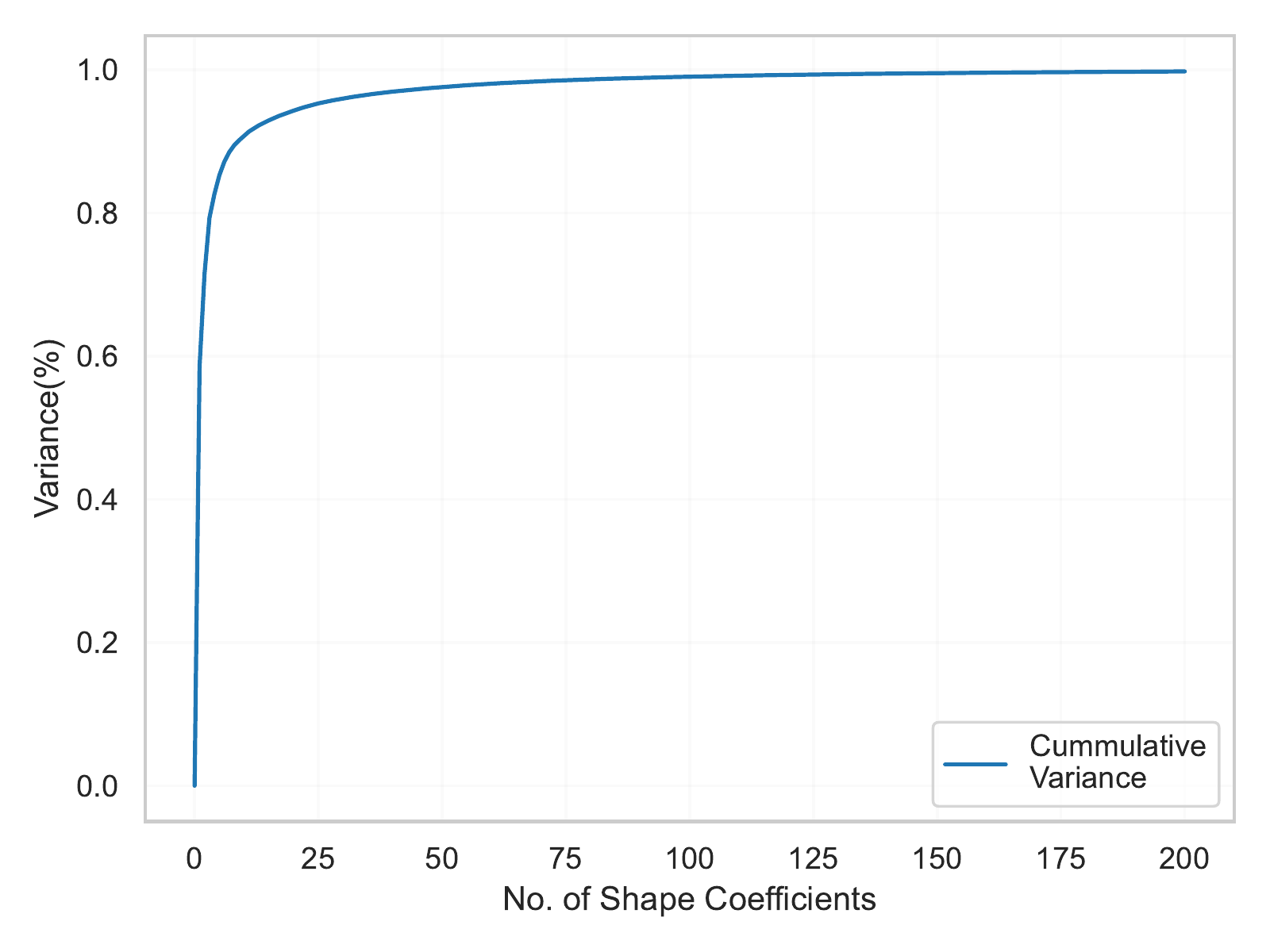}\label{fig:pca_res}
\caption{Cumulative variance of the skin-bone-organ model shape space.}
\label{fig:pca_cum}
\end{figure}

\begin{equation}\label{eq:model_params_3}
\underset{\vecb{\Tilde{\theta}}_{bo}^i,\vecb{\bar{\theta}}_{bo}^i,\mathcal{J}_{bo}, \mathcal{W}_{bo}}{\arg \min } \sum_{i}^{S_n} || \mathcal{P}(\mathbf{\bar{U}}_{bo}^i, \mathbf{U}_{s}^i) -  \mathcal{P}(\mathbf{\bar{V}}_{bo}^i, \mathbf{V}_{s}^i) ||
\end{equation}

\begin{figure*}[t]
	\centering
	\begin{subfigure}[b]{0.45\textwidth}
	\centering	\includegraphics[width=\textwidth,keepaspectratio]{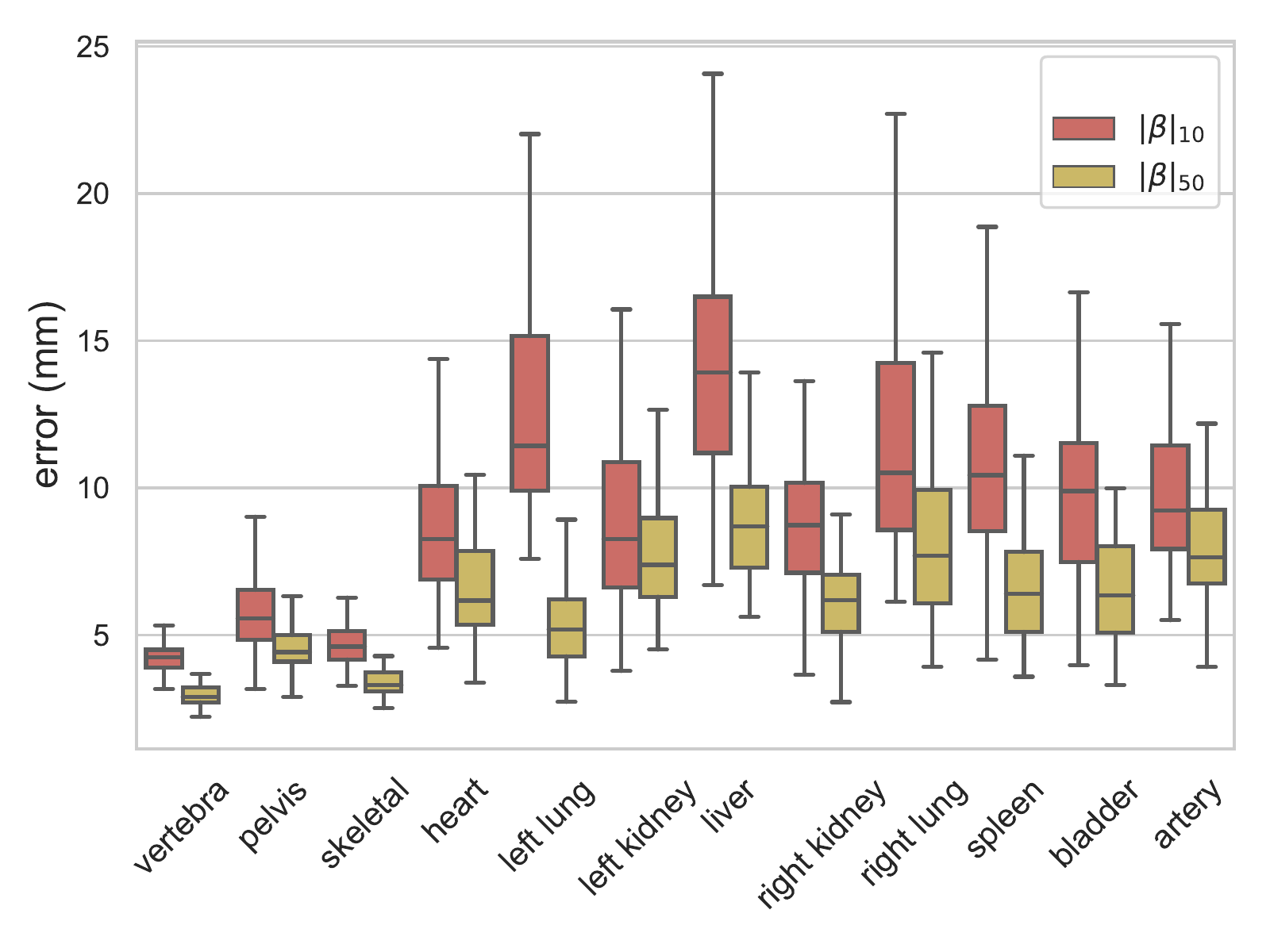}	
    \caption{}\label{fig:acrin_1}
    \end{subfigure}
    	\begin{subfigure}[b]{0.45\textwidth}
	\centering
	    \includegraphics[width=\textwidth]{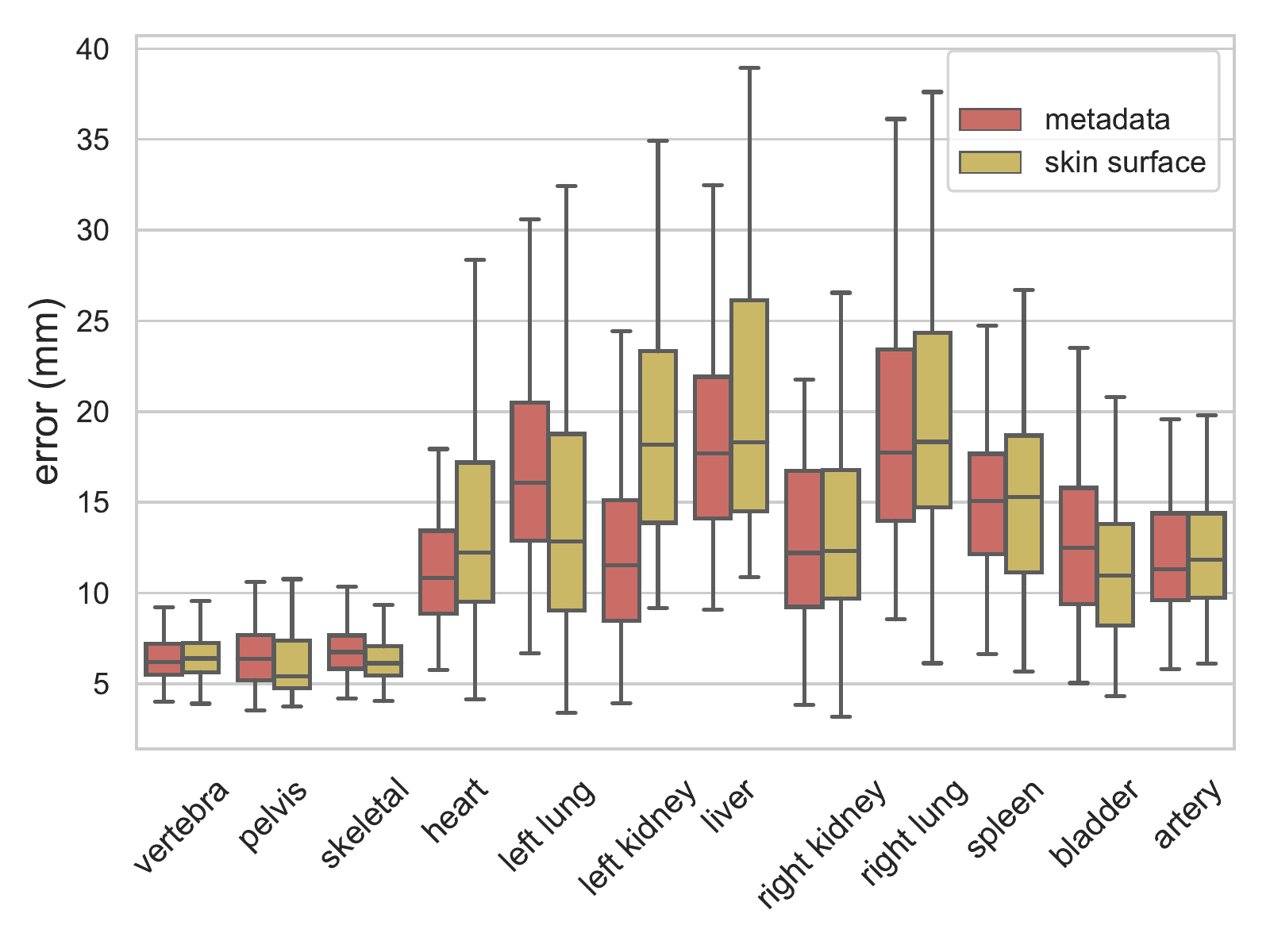}
    \caption{} \label{fig:acrin_2b}
    \end{subfigure}
 	\caption{(a) Box-plot of model generalization on the registered ACRIN dataset using $10$ and $50$ shape components. (b) Box-plot of model generalization on the registered ACRIN dataset by only using the metadata and skin surface as a guide to determine the internal body shape.}

\end{figure*}

\begin{figure}[h]
	\centering
    \includegraphics[width=\linewidth]{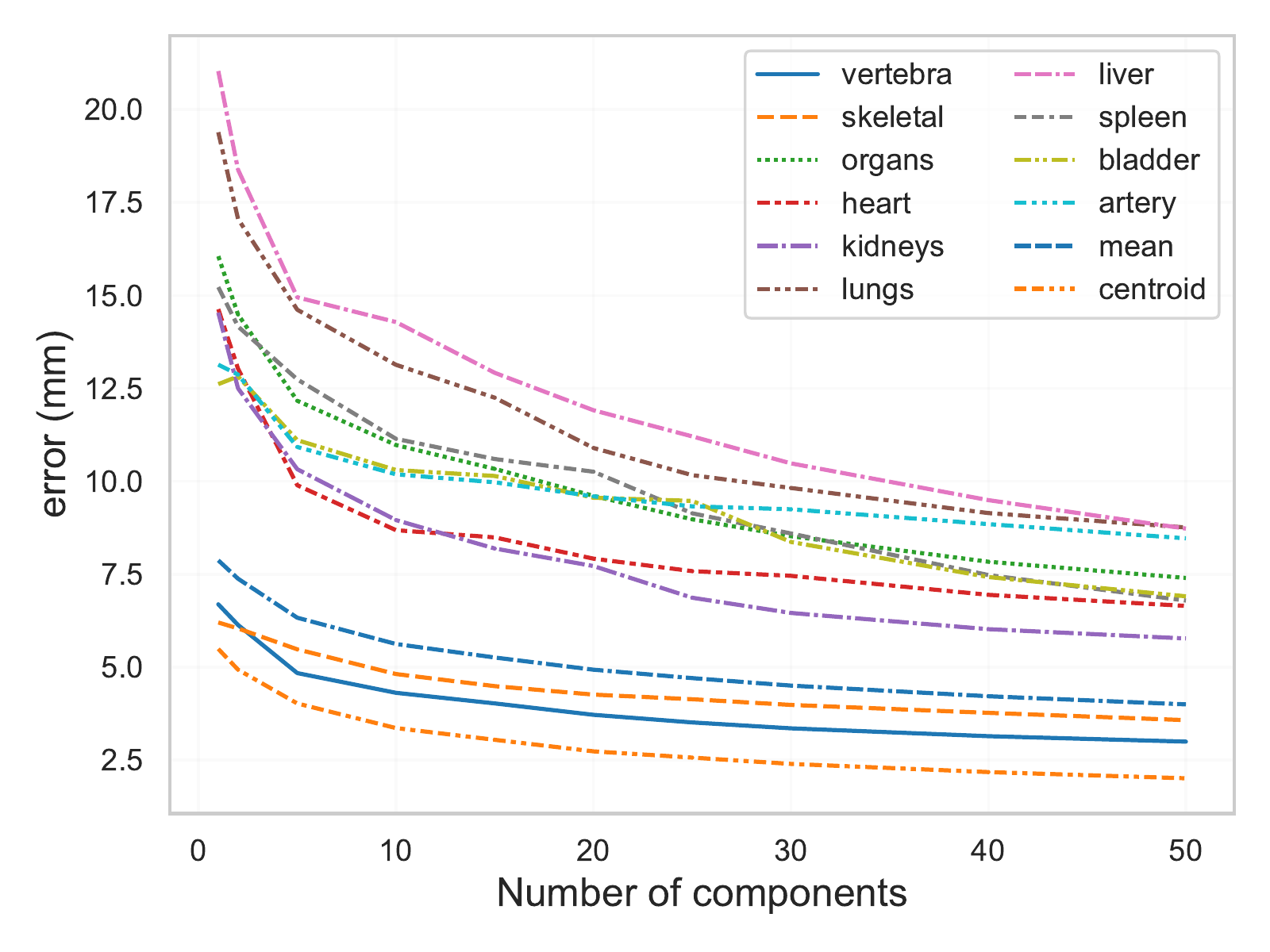}   
    \caption{Residual error on ACRIN dataset with respect to the different number of shape components. The centroid represents, the mean error of the centre of mass for individual segments between reconstructed and registered models.}\label{fig:acrin_2}

\end{figure}

We start off by initialising the joint regressor $\mathcal{J}_{bo}$ on the template bone-organ model $\mathbf{T}_{bo}$ based on the joint locations from its initial kinematic chain. The joints are always located between $2$ bone segments. We randomly sample 50 closest vertices to the joint from both segments where the vertex normals approximately faces the vertex-joint direction. The joint regressor $\mathcal{J}_{bo}$ is learnt using a least square fit for the sampled vertices. The rest pose vertices $\mathbf{U}^{i}_{bo}$ are initialized $M^{bo}(\vecb{\hat{\beta}}^i_b)$ with  only the scale parameters $\vecb{\hat{\beta}}^i_b$ obtained from the initial rigid bone fit.  Similar to the skin model, the blend weights $\mathcal{W}_{bo}$ and poses $\vecb{\Tilde{\theta}}_{bo}^i$ are initialized.
During unposing of the bone-organ model, we define that the motion of ribs, sternum and pelvis are a function of shape rather than a function of the pose. Hence, we initialize the pose to zero for these particular segments.  Note that, only the ribs and sternum are leaf nodes in our kinematic chain, i.e. there are no child segments. However, for the femur, we additionally include the pelvis rotation, as it is its parent node.  

\begin{figure}[!hb]
	\centering
    \includegraphics[width=1\linewidth,keepaspectratio]{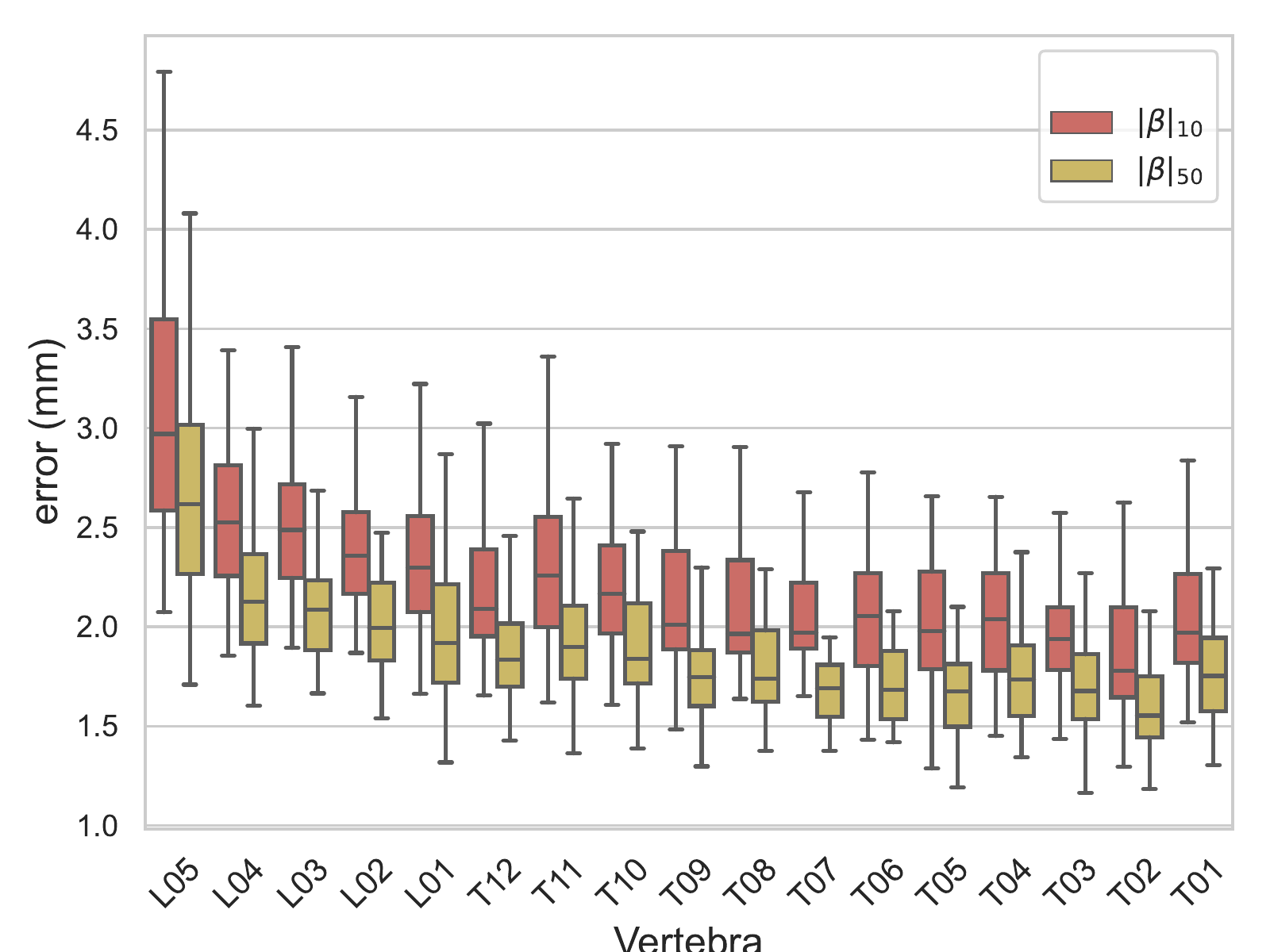}
    \caption{Model generalization on VerSe dataset using $10$ and $50$ shape components. }
       \label{fig:generalization_verse}
\end{figure}

To stabilize the optimization we make use of similar regularizes defined in the skin model for both the objectives functions. For the edge loss, we additionally incorporate the virtual edges from the registration process with lower weightage to reduce mesh interpenetration. Symmetry regularizer is applied only to the bone structures. Regularizing the joint locations of bone and skin along the arms and legs are done by minimizing the following equation:

\begin{equation*}
|| \mathcal{J}_{bo}(\mathbf{U}_{bo}^i) - \mathcal{J}_{s}(\mathbf{U}_{s}^i)||^2.
\end{equation*}

\begin{figure*}[h]
	\centering
	\begin{subfigure}[b]{0.32\textwidth}
	\centering
	    \includegraphics[width=\textwidth,keepaspectratio]{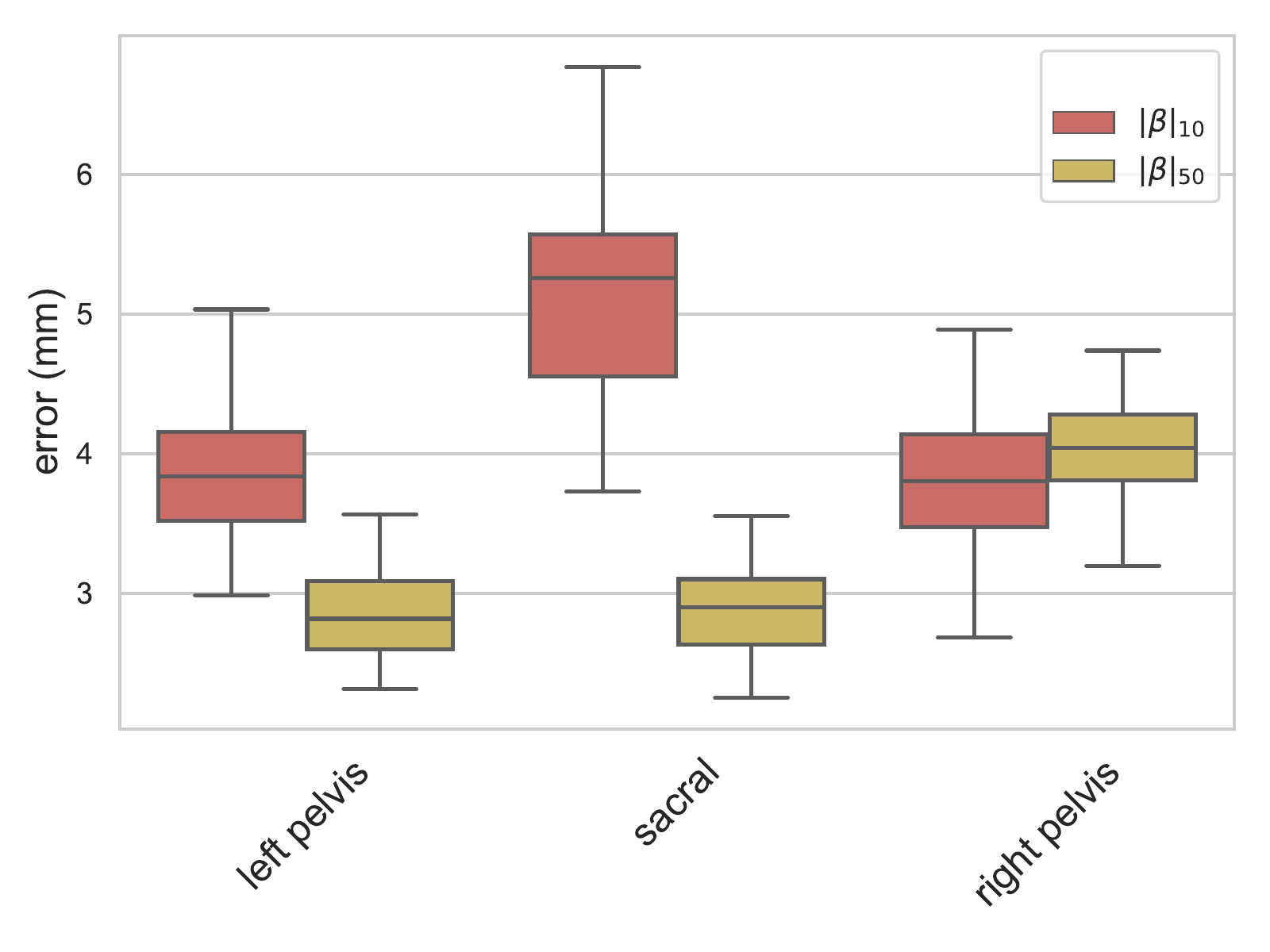}
    \caption{CTPEL}
    \end{subfigure}
    	\begin{subfigure}[b]{0.32\textwidth}
	\centering
	    \includegraphics[width=\textwidth]{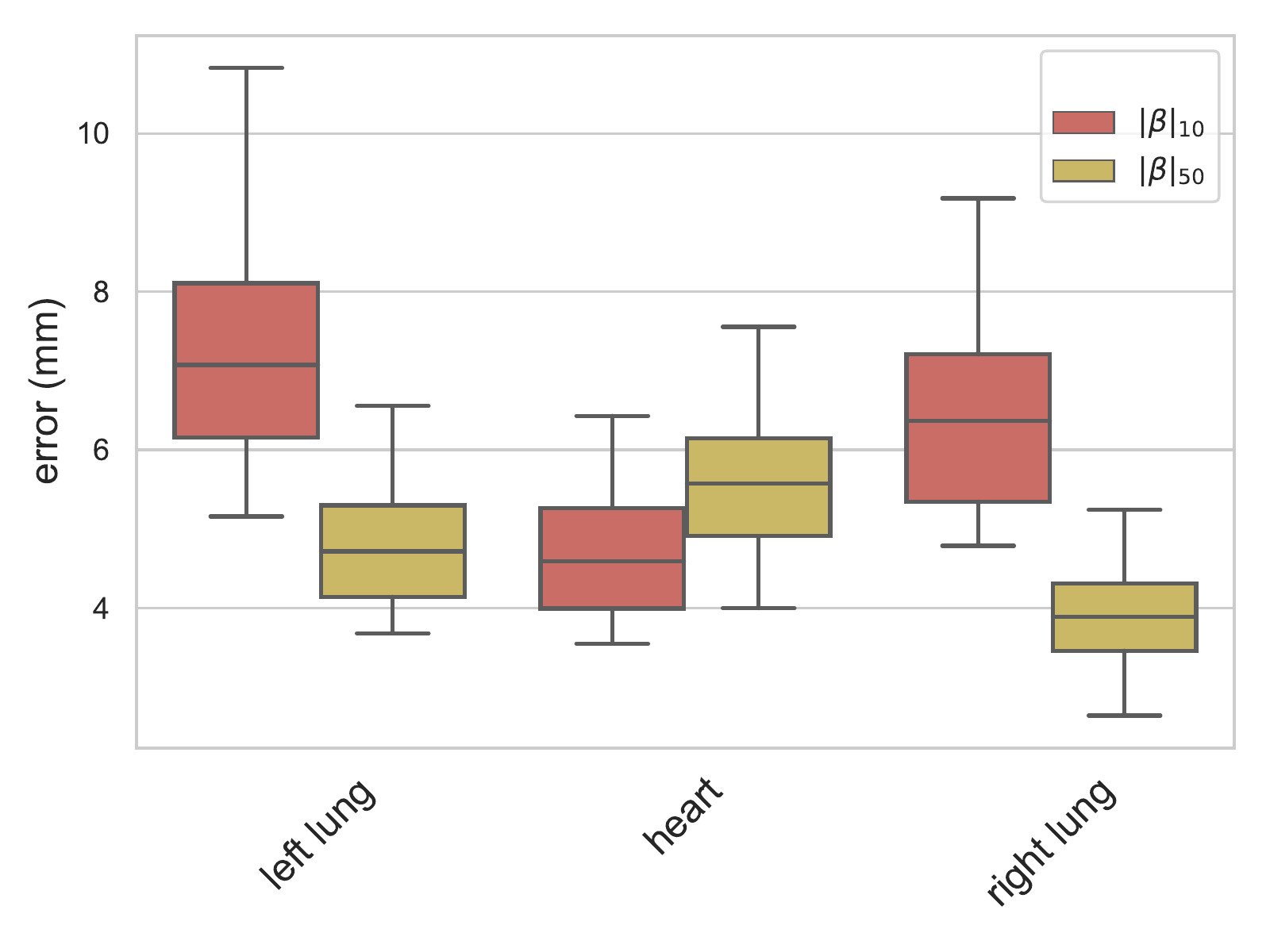}
    \caption{SturctSeg}
    \end{subfigure}
    \centering
	\begin{subfigure}[b]{0.32\textwidth}
	\centering
	    \includegraphics[width=\textwidth,keepaspectratio]{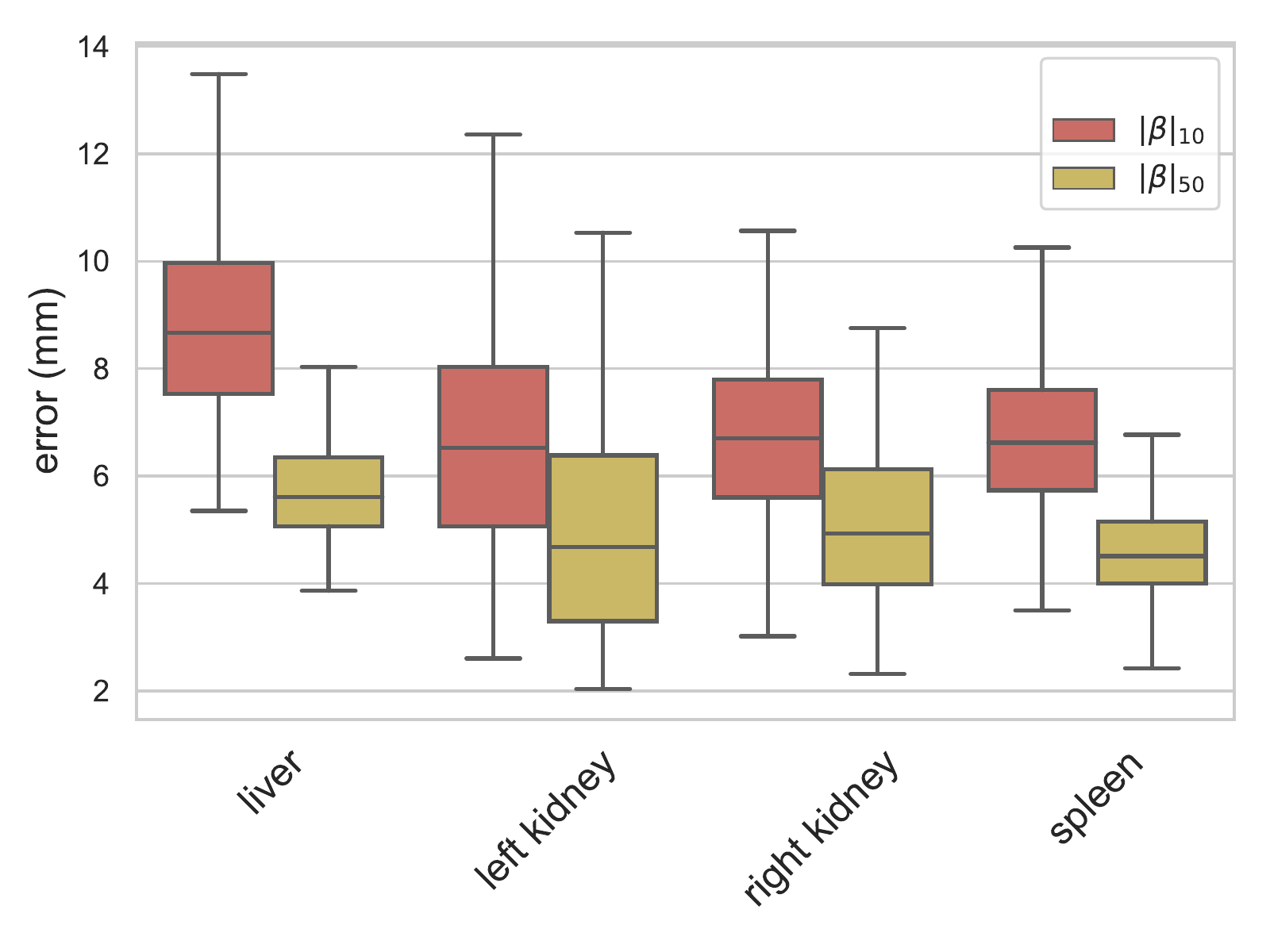}
    \caption{Abdomen1k}
    \end{subfigure}

 	\caption{	Model generalization on various datasets using $10$ and $50$ shape components.}
   \label{fig:generalization}
\end{figure*}

We alternate between optimizing Eq.~\ref{eq:model_params_2} and Eq.~\ref{eq:model_params_3}, while carrying the optimized parameters between them. For  optimization of Eq.~\ref{eq:model_params_3}, we initialize  the pose $\vecb{\bar{\theta}}_{bo}^i$ with zero, and regularize them towards zero. While alternating to optimize Eq.~\ref{eq:model_params_2}, we initialize unposed vertices $\mathbf{U}_{bo}^i$ with the obtained vertices $\mathbf{\bar{U}}_{bo}^i$ after optimizing  Eq.~\ref{eq:model_params_3}. 

\subsubsection{Shape Space}
From the unposed skin $\mathbf{U}_{s}^i$ and bone-organ $\mathbf{U}_{bo}^i$ volume, we learn the shape components with the aid of mean and principal shape components. We do not have complete registrations around skulls, arms and legs for some of the volume. Hence, we use a publicly available implementation~\footnote{https://github.com/allentran/pca-magic} of Probabilistic Principal Component Analysis (PPCA)~\cite{ppca}, which can handle missing data. By performing PPCA, we obtain a mean skin $\mathbf{T}_\mu^s$, bone-organ $\mathbf{T}_\mu^{bo}$ and vertex offsets to the mean in the form of shape space for skin $\mathbf{B}_\mu^s$ and $\mathbf{B}_\mu^{bo}$ bone-organ.

\section{Evaluation}
\begin{figure*}[ht]
\centering
\includegraphics[width=0.8\textwidth,keepaspectratio]{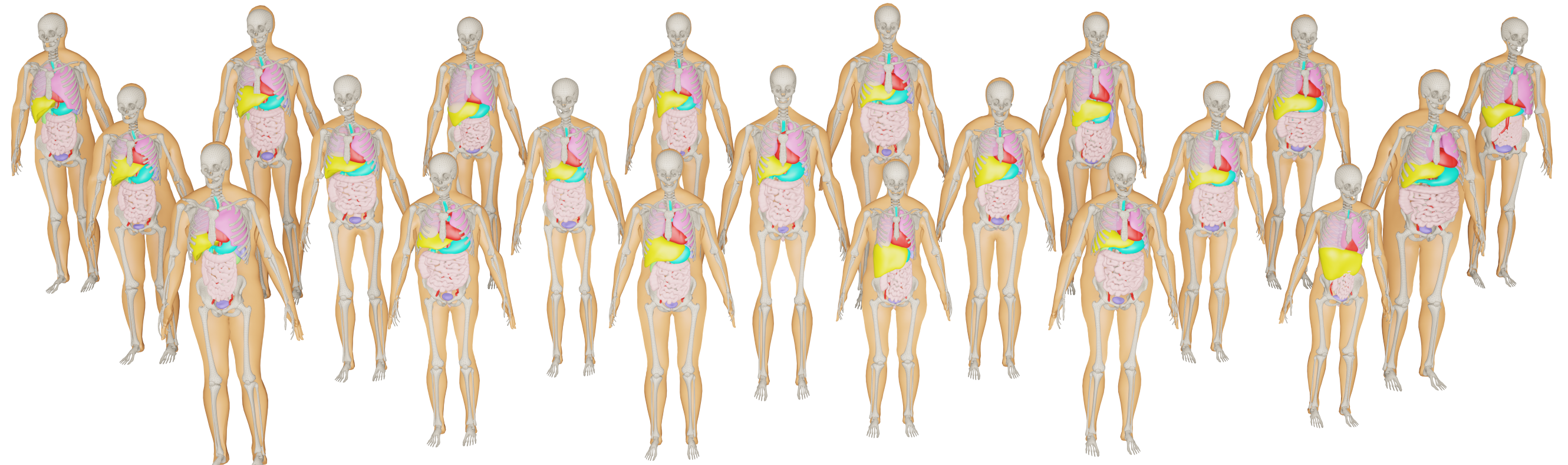}
\caption{Random samples drawn from the  model. }\label{fig:random_sample}
\end{figure*}

In Fig.~\ref{fig:pca}, we visualize the first three shape components, while Fig.~\ref{fig:pca_cum} displays the cumulative variance of the full model. The first 10 shape components captures $88\%$ of the variance, and the first 20 components capture $92\%$ of the variance. Although the shape space of the skin and bone-organ are coupled, the kinematic model is separated to account for differences in skeletal posture. We use a neutral model rather than separating into two genders on basis of the number of CT volumes available. We map the gender, height, and weight of the registered patients where available to the shape space coefficients using  a linear regressor. We provide qualitative results of random samples generated in Fig.~\ref{fig:random_sample}.

To demonstrate the effectiveness of the skin-bone-organ model we perform evaluations on multiple datasets. First we assess the model generalizability on a registered test set. We then demonstrate model completion from metadata such as height, weight, or surface scans. We finally evaluate the model performance on various public datasets, where segmentations of organs or bones are provided.

\subsection{Model Generalization}\label{mg}
We register the skin, bones and organs on separate held out test set. For the test set we use CT volumes from ACRIN-NSCLC-FDG-PET~\cite{acrin}. We ignore all volumes where the z-spacing is greater then $3.27$ mm, while retaining volumes greater than $2.5$ mm only if clear separation of the vertebra is visible. In total we test on 78 scans, which were registered to its segmentations as described in the methodology section. 

We minimize the vertex-to-vertex error between the reconstructed model and the registration while regularizing the process with shape and pose priors. The shape prior keeps the reconstructed model close to its mean, while the pose prior restricts individual vertebrae from deviating too much from their parent and child nodes. It's important to note that we do not translate individual segments, and all the shape variations are accounted for only by the shape space of the model. In the case of the ACRIN dataset, regions such as the legs, arms, and part of the skull are not available, so we exclude these regions while measuring metrics and during reconstruction.

In Figure~\ref{fig:acrin_1}, we present the box plot illustrating the error in individual segments for $10$ and $50$ shape components. The vertebra includes all vertebral sections, the pelvis includes the pelvis and sacral, and the skeleton comprises the remaining bone structures. In Figure~\ref{fig:acrin_2}, we display the generalization error across different numbers of shape components. The results indicate that the bones can be generalized well, with an error of $3.66$ mm at $25$ components. However, the errors for individual organs are significantly higher, as we optimize the model for the entire body rather than individual organs. This is expected since the shape space accounts for variations in organ shapes and locations. Using only $25$ components, we achieve an average organ error of $8.83$ mm, which can provide a starting point for downstream tasks involving organ shape and placement. We also measure the mean centroid error for each segment, which represents the center of mass error. With $25$ components, we observe a mean centroid error of  $2.52$ mm. 

Fig.~\ref{fig:acrin_2b} displays the ability of our model to estimate the shape of inner organs and skeleton using patient metadata or skin surface. For estimating with metadata, we use the linear regressor with patient's gender, weight, and height to determine the model shape coefficients. On the other hand, when using the skin surface, we only use the shape coefficients of our skin model that fit the target skin. In both cases, we optimize only for the pose while using the previously determined shape coefficients. We observe an average overall error of $8.11$ mm and $8.68$ mm using metadata and skin surface, respectively.

\subsection{Model Generalization on Public Organ Segmentations}
We evaluate the generalization of our model on public datasets containing multiple organs or bones. Specifically, we test our model on Verse~\cite{verse} for vertebra, CTPEL~\cite{ctpel} for pelvis and sacral, Abdomen1K~\cite{abd1k} for liver, kidney, and spleen, and StructSeg~\cite{structseg} for lungs and heart. For all methods, we follow a similar registration process. To register to each dataset, we obtain a surface mesh of the segmented volume using marching cubes~\cite{marchingcubes}, and then minimize the chamfer loss between the vertices of our model and the target surface mesh.
For regularization we use the same procedure as defined in Sec.~\ref{mg} by using pose and shape prior with no individual translations, i.e. the combined shape model has to best represent all target surfaces. The final errors presented are the bidirectional point-to-surface distance~\cite{spine_model}. 

The model's generalization performance for the vertebrae region on the Verse dataset is presented in Fig.\ref{fig:generalization_verse}. The full Verse dataset, including the train, test and validation sets, was used for testing. However, volumes containing L6 and T13 vertebrae, which are not represented by our model, were excluded.  The generalization errors for the pelvis and sacral region in CTPEL, for the liver, kidney, and spleen in StructSeg, and for the lungs and heart of the Abdomen1k dataset are presented in Fig.~\ref{fig:generalization}. In the Abdomen1k dataset, volumes containing incomplete liver were excluded from the analysis. The results obtained on these datasets are consistent with those observed on the ACRIN dataset.

\section{Conclusion}
 A digital twin of a patient can be a valuable tool to improve multiple clinical tasks, such as automating clinical workflows, estimating patient-specific X-ray dose exposure to patient and medical staff, positioning, markerless tracking and navigation assistance in image-guided
interventions. For the first time, we provide a joint model with an accurate estimation of shapes
and poses of the patient surface, skeleton and internal organs.  We propose a deformable shape and pose human model devised out of individual segments such as the skin, internal organs and bones, learnt from surfaces extracted from segmented whole-body CT images. 
With surface errors of 3.66 mm for bones and an average organ error of 8.83 mm, we believe that the statistical model will allow many automation tasks in clinical workflows and lays the basis for fast intraoperative model personalisation given patient-specific information.

\section*{Disclaimer}
The concepts and information presented in this article are based on research and are not commercially available.

\bibliographystyle{IEEEtran}
\bibliography{refs}
\end{document}